\documentclass[10pt,twocolumn,letterpaper]{article}

\usepackage[pagenumbers]{cvpr} 

\usepackage{graphicx}
\usepackage{amsmath}
\usepackage{amssymb}
\usepackage{booktabs}
\usepackage{kotex}
\usepackage{bbm}
\usepackage{multirow}
\DeclareMathOperator*{\argmax}{arg\,max}

\newcommand{\norm}[1]{\left\lVert#1\right\rVert}

\usepackage[pagebackref,breaklinks,colorlinks]{hyperref}

\usepackage[capitalize]{cleveref}
\crefname{section}{Sec.}{Secs.}
\Crefname{section}{Section}{Sections}
\Crefname{table}{Table}{Tables}
\crefname{table}{Tab.}{Tabs.}

\def\cvprPaperID{694} 
\def\confName{CVPR}
\def\confYear{2022}

\begin{document}

\title{
Exploiting Inter-pixel Correlations in Unsupervised Domain Adaptation \\for Semantic Segmentation 
}

\author{Inseop Chung\quad Jayeon Yoo\quad Nojun Kwak\\
Seoul National University\\
{\tt\small \{jis3613,jayeon.yoo,nojunk\}@snu.ac.kr}
}
\maketitle

\begin{abstract}
    ``Self-training" has become a dominant method for semantic segmentation via unsupervised domain adaptation (UDA). It creates a set of pseudo labels for the target domain to give explicit supervision. However, the pseudo labels are noisy, sparse and do not provide any information about inter-pixel correlations. We regard inter-pixel correlation quite important because semantic segmentation is a task of predicting highly structured pixel-level outputs. Therefore, in this paper, we propose a method of transferring the inter-pixel correlations from the source domain to the target domain via a self-attention module. The module takes the prediction of the segmentation network as an input and creates a self-attended prediction that correlates similar pixels. The module is trained only on the source domain to learn the domain-invariant inter-pixel correlations, then later, it is used to train the segmentation network on the target domain. The network learns not only from the pseudo labels but also by following the output of the self-attention module which provides additional knowledge about the inter-pixel correlations. Through extensive experiments, we show that our method significantly improves the performance on two standard UDA benchmarks and also can be combined with recent state-of-the-art method to achieve better performance.
\end{abstract}
\vspace{-5mm}

\section{Introduction}
\label{intro}
Unsupervised domain adaptation (UDA) has become one of the primary methods to tackle label-scarce real-world problems such as simulated training for robots and autonomous driving. It transfers the knowledge of a deep neural network learned from the source domain with rich labels to the target domain without any label information. Usually, the source domain is photo-realistic synthetic dataset \cite{Richter_2016_ECCV, RosCVPR16} and the target domain is real world images \cite{Cordts2016Cityscapes}. UDA is especially useful for semantic segmentation since it requires dense pixel-level labeling process which is laborious and time-consuming. Recently, the ``self-training" scheme has become a dominant method to enable UDA for semantic segmentation. This paper also tackles the UDA for semantic segmentation based on the ``self-training" scheme, yet we introduce a novel method for transferring inter-pixel correlations in the source domain to the target domain, which is also simple and effective.

The idea of self-training \cite{zou2018unsupervised,zou2019confidence, li2019bidirectional} is to give supervision for the unlabeled target domain as well. It generates a set of pseudo labels for the target domain using a trained model and then retrains a new model with the generated pseudo labels. To enhance the performance, it iteratively repeats this process with the retrained model. However, since pseudo labels are based on confident pixels whose predicted confidence surpasses a certain threshold, they are not always correct and sometimes very sparse which means that some pixels have no pseudo labels. To resolve this issue, recent works \cite{shin2020two,zhang2021prototypical,zheng2021rectifying} have proposed to generate better pseudo labels. They denoise, rectify, and densify the pseudo labels to give more correct supervision. However, these methods are somewhat heuristic and require complex algorithms. Also, pseudo labels provide information only about which class a pixel should be predicted as but not about how each pixel should be correlated or similar to other pixels. We argue that providing guidance about the correlation between pixels is important since the outputs of a semantic segmentation task are highly structured and correlated. 

A semantic segmentation task requires a model to predict highly structured pixel-level outputs which contains densely related spatial and local information. For example, the pixels corresponding to the \textit{sky} are usually located at an upper area of an image, \textit{car} pixels are over the \textit{road} pixels and \textit{sidewalks} and \textit{fence} pixels are usually on the sides of the \textit{road}. These inter-pixel relations are domain-invariant properties which are consistent for different domains. Therefore, in this paper, we propose an idea of transferring this domain-invariant knowledge of the inter-pixel correlation from the source domain to the target domain. 
Our intuition is that providing the extra knowledge of domain-invariant pixel correlations can help to overcome and complement the defects of sparse and incorrect pseudo labels.

To effectively capture these inter-pixel correlations and transfer them to the target domain, we propose a novel self-attention \cite{cheng2016long, parikh2016decomposable, vaswani2017attention} module. The proposed self-attention module takes predictions of the segmentation network and outputs new self-attended predictions. The module creates an attention map which captures similarities between pixels and uses this map to calculate a weighted sum of the given predictions. Our proposed algorithm consists of two steps. The first step is to train the self-attention module only on the source domain since it has the ground truth labels that can correctly guide about the domain-invariant inter-pixel correlations. The second step is to train the actual segmentation network using the trained self-attention module by transferring its learned knowledge to the target domain. At this phase, the attention module is frozen and the main segmentation network is the only part being trained. The module takes the predictions of the segmentation network as inputs and outputs new self-attended predictions. Then, we train the segmentation network to follow the the new self-attended predictions. We call this the `self-attention loss'. Overall, the segmentation network is trained with both the self-attention loss and the self-training loss. Detailed training methodology will be explained in Sec.~\ref{method}. 

We conduct extensive experiments on the two standard UDA benchmarks, GTA5\textrightarrow Cityscapes and SYNTHIA\textrightarrow Cityscapes. We show that the network trained with our method outperforms the one trained only with self-training. Especially, our method shows significant improvements on rare classes which scarcely appear in the datasets. Also, our method does not require a complex human-designed algorithm, it rather lets the module to decide which knowledge to be learned and transferred, thus our method is simple and needs less human intervention. Moreover, our method can be applied to other state-of-the-art self-training UDA methods and additionally boost the performance. We combined our method with ProDA \cite{zhang2021prototypical}, one of the recent SOTA methods, and could observe a performance improvement which sets new SOTA scores on both UDA benchmarks tested.

\begin{figure*}[t]
    \centering
    \includegraphics[width = 1.0\linewidth]{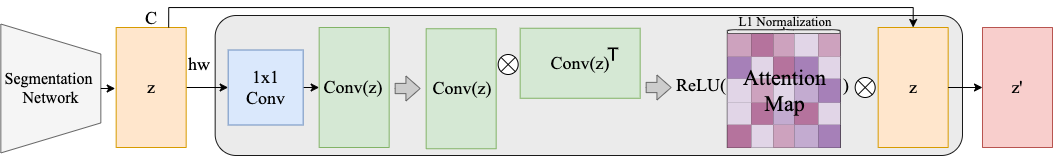}
    \vspace{-5mm}
    \caption{The architecture of the proposed self-attention module. Grey shaded area at the center is the self-attention module and $\bigotimes$ refers to a matrix product. The module takes the output of the segmentation network and creates an attention map to apply self-attention to $z$. $z'$ is the new self-attended output of the module which possesses information about inter-pixel correlations.}
    \vspace{-3mm}
    \label{fig:module_arch}
\end{figure*}

\section{Related Work}
\subsection{Unsupervised Domain Adaptation}
Unsupervised domain adaptation has been widely studied to address the domain gap between labeled source data and unlabeled target data. 
Since domain adaptation is an important problem in real world environments, it has been studied in various tasks such as classification\cite{ganin2018dann} and segmentation\cite{zou2018unsupervised, zou2019confidence, zhang2021prototypical}. There are three major lines of domain adaptation approaches: adversarial learning, image translation, and self-training. Based on the theoretical analysis\cite{ganin2018dann}, the works \cite{tsai2018learning, vu2019advent} adapt the segmentation model to the target domain via adversarial learning which tries to increase the domain discrimination loss of the discriminator while reducing the segmentation loss on the source domain. The works \cite{hoffman2018cycada, yang2020fda} translate the labeled source domain images into the target domain styles using CycleGAN~\cite{CycleGAN2017} and Fourier transforms to encourage image-level adaptation.

\subsection{Self-training}
Another line of unsupervised domain adaptation is self-training which is first introduced in semantic segmentation by \cite{zou2018unsupervised}. In \cite{zou2018unsupervised}, the authors generate pseudo labels using a model trained on the source domain and retrain the model with the pseudo labeled target domain images to implicitly encourage class-wise feature alignment and enable adaptation to the target domain. However, since pseudo labels are noisy due to domain discrepancy, many methods have been proposed to address this problem. \cite{zou2019confidence} introduces a regularization method to generate soft pseudo labels and smooth the output of the model to avoid overfitting on overconfident labels. \cite{zhang2021prototypical} and \cite{zheng2021rectifying} lower the weights for noisy pseudo labels based on the feature distance from the class prototypes and the variance between the outputs of the two segmentation models respectively. In addition, \cite{guo2021meatcorrection} has the same effect of modifying pseudo labels based on prior distribution by introducing a transition probability map between the ground-truth label and the pseudo label. To address the sparsity of confident pseudo labels which is another limitation of self-training, \cite{shin2020two} proposes densification of pseudo labels using confident pseudo labels of neighboring pixels. In this paper, we resolve these issues by transferring inter-pixel correlation which is a domain-invariant property.

\subsection{Self-attention for vision}
Self-attention calculates the response at a position in a sequence (\emph{e.g.}, an image) by interacting with all positions and paying more attention to more salient positions. There have been several attempts to introduce attention modules into convolutional networks. Non-local network \cite{wang2018nonlocal} captures long-range dependencies by computing interactions between any two positions regardless of their positional distance. Another study \cite{woo2018cbam} refines features by using channel and spatial attention for various vision tasks. Recently, transformer\cite{vaswani2017transformer} based models, which consist of attention mechanisms rather than convolutional architectures, are also being actively studied in vision tasks \cite{carion2020detr, dosovitskiy2021ViT}.

\section{Preliminary}
\label{preliminary}
In the common UDA setting, it is assumed that the dataset in the source domain has labels, $\{x_s, y_s\}$, while that in the target domain is unlabelled, $\{x_t\}$. The main purpose of UDA for semantic segmentation is to train a segmentation network, $\mathcal{G}$, to perform well on the target domain by transferring the knowledge it has learned from the source domain. The two domains are assumed to share the same $C$ classes, hence it is possible to adapt between them. Typically, for the source domain, the categorical cross-entropy loss is used to train the network utilizing the given ground truth labels:
\vspace{-2mm}
\begin{equation}
    \mathcal{L}_{seg}^S(x_s) = -\sum_{i=1}^{H\times W} \sum_{c=1}^C y_s^{(i,c)}\log (p_s^{(i,c)}).
    \label{eq:seg_src}
    \vspace{-2mm}
\end{equation}
Here, $p_s = \mathcal{G}(x_s)$ and $p_s^{(i,c)}$ refers to the softmax probability $p_s^{(i)}$ of pixel $i$ belonging to class $c$. 
The label $y_s^{(i)}$ is a one-hot vector where $y_s^{(i,c)}$ is 1 if the ground truth class at pixel $i$ is $c$, otherwise it is 0.
However, for the target domain, the labels do not exist, thus pseudo labels $\{\hat{y}_t\}$ are generated to perform self-training.
The pseudo labels are based on the confident pixels predicted by a trained network.
Following the process of \cite{li2019bidirectional}, the pseudo labels are assigned only to pixels with confidence higher than a given threshold:
\vspace{-1mm}
\begin{equation}
    \hat{y}_t^{(i,c)} =
    \begin{cases}
    1, & \text{if }\ c = \argmax_{c'} p_t^{(i,c')}  \\
       & \text{and}\ \max_{c'} p_t^{(i,c')} > \tau^{c} 
    \\
    0, & \text{otherwise}
    \end{cases}
    \vspace{-1mm}
\end{equation}
\begin{equation}
    \mathcal{L}_{seg}^T(x_t) = -\sum_{i=1}^{H\times W} \sum_{c=1}^C \hat{y}_t^{(i,c)}\log (p_t^{(i,c)}).
    \label{eq:seg_trg}
    \vspace{-1mm}
\end{equation}
The threshold $\tau^{c}$ is set differently for each class $c$ and how it is determined is detailed in the supplementary.
Using the generated pseudo labels, the target domain can also be trained in a supervised fashion as well using (\ref{eq:seg_trg}). The overall segmentation loss is the sum of (\ref{eq:seg_src}) and (\ref{eq:seg_trg}). In this way, the performance of the network on the target domain can be greatly improved, however, in this work, we further enhance this by introducing our novel self-attention module.

\begin{figure*}[t]
    \centering
    \includegraphics[width = 0.9\linewidth]{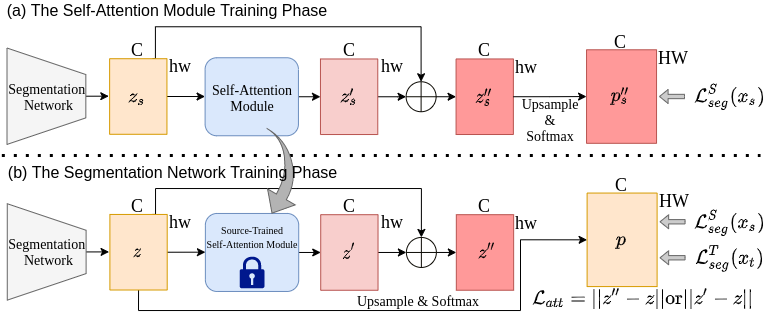}
    \vspace{-5mm}
    \caption{Training of the self-attention module and the segmentation network. (a) The module is trained only on the source domain to learn the correct pixel relations via the ground truth labels of the source domain. (b) The source-trained self-attention module is used for the domain adaptive training of the segmentation network. The segmentation network is trained by making $z$ to follow the output of the module, $z'$ or $z''$ along with self-training scheme. $\mathcal{L}_{att}$ can be computed for both domains.}
    \vspace{-5mm}
    \label{fig:Trainig}
\end{figure*}

\section{Method}
\label{method}

\subsection{Self-attention Module (SAM)}
A segmentation network uses an upsampling to enlarge its output to the size of its input. DeepLabV2 \cite{chen2017deeplab} uses bilinear interpolation while FCN-8s \cite{long2015fully} uses transposed convolution. Therefore, it can be formulated as $p = S(U(z)) = \mathcal{G}(x) \in \mathbb{R}^{H\times W\times C}$ where $S$ and $U$ stands for the softmax and the upsampling respectively. The $z \in \mathbb{R}^{h\times w\times C}$ is the output of the network right before the upsampling ($h,w < H,W$). It is the logit which becomes the class probability distribution after softmax is applied. As shown in Fig.~\ref{fig:module_arch}, $z$ is inputted to our self-attention module to create a new self-attended output $z'$. We assume that $z$ is flattened in the spatial dimension for simplicity, so its shape is $\mathbb{R}^{hw\times C}$. Then, $z$ consists of $hw$ number of logits, $z=(z_1, z_2, z_3, \dots z_{hw})^\mathsf{T}$ where each logit, $z_j$, is a $C$-dimensional vector. The architecture of the self-attention module is depicted in Fig.~\ref{fig:module_arch}. First, $z$ is transformed into another space using a $1\times1$ convolutional layer producing Conv($z$) $\in \mathbb{R}^{hw\times C}$. Then we calculate the attention map by outer-producting Conv($z$) to itself. More specifically, if we denote Conv($z$) as $A$ to keep the notation uncluttered, the attention map is computed as 
\vspace{-2mm}
\begin{equation}
    M = \frac{{A}\boldsymbol{\cdot} A^\mathsf{T}}{\norm{A}_2 \boldsymbol{\cdot} \norm{A}^\mathsf{T}_2} \in \mathbb{R}^{hw\times hw},
    \vspace{-2mm}
\end{equation}
where $\boldsymbol{\cdot}$ is a matrix product, $\norm{A}_2 \in \mathbb{R}^{hw\times 1}$ refers to the row-wise $L2$-norm of $A$ along with its channel dimension and the division is done element-wise.
Each row of $M$ shows how each pixel is similar to other pixels in cosine similarity, hence its element is normalized between $-1$ and 1. We apply ReLU~\cite{agarap2018deep} activation on $M$ to eliminate any negative similarity and then $L1$ normalize each row in order to make the sum of all elements in the same row equals 1. 
We denote $M'$ as the attention map after the ReLU activation and the normalization. $M'_{[i,j]}$ refers to an element of $i$th row and $j$th column of $M'$ and is computed as follows:
\vspace{-2mm}
\begin{equation}
    M'_{[i,j]} = \frac{ReLU(M)_{[i,j]}}{\sum_{k=1}^{hw} ReLU(M)_{[i,k]}}
    \vspace{-2mm}
\end{equation}
The last step is to matrix product $M'$ and the input $z$ to produce the final output $z'$:
\vspace{-2mm}
\begin{equation}
    z' = M'\boldsymbol{\cdot}z \in \mathbb{R}^{hw\times C}
    \vspace{-2mm}
\end{equation}
Now $z'$ is the new self-attended output attended by $M'$. 
Each logit (row) of $z'$ is a weighted sum of the logits of $z$ whose weights are the corresponding row of $M'$:
\vspace{-2mm}
\begin{equation}
    z'_i = \sum_{j=1}^{hw} M'_{[i,j]}z_j.
    \vspace{-2mm}
\end{equation}
Here, $z'_i$ and $z_j$ refer to a logit (row) of $z'$ and $z$ respectively. 
Therefore, $z'$ is a set of self-attended logits that captures the information about the inter-pixel correlations based on the similarity between pixels informed by the attention map $M'$. Note that we use $z$ instead of $U(z)$ as our input to the module because using $U(z)$ requires expensive computation and huge memory when computing $M$. 

Fig.~\ref{fig:Trainig} (a) shows the training of the self-attention module. As mentioned earlier, we train the self-attention module \textit{only on the source domain} because we want the module to learn the domain-invariant inter-pixel correlations and the source domain has the ground truth labels that can guide this correctly. We add the output of the segmentation network $z_s$ to the output of the module $z'_s$ to create $z''_s = z_s + z'_s$. We upsample and softmax $z''_s$ to produce $p''_s = S(U(z''_s))$ and compute the segmentation loss (\ref{eq:seg_src}) using it. The segmentation network and the self-attention module are jointly trained by minimizing the source segmentation loss. However, when the training is finished, the trained segmentation network is discarded and only the self-attention module is saved to be used later for the next step which is the actual domain adaptive training of the segmentation network. 

When computing the source segmentation loss (\ref{eq:seg_src}), $z''_s$ is used instead of $z'_s$ to make a skip connection which enables the loss signal to directly flow to the segmentation network during back-propagation. If there were no skip connection and the loss were computed on $z'_s$, the segmentation network would have been trained to produce output that only performs well after the attention module is applied, eventually producing inaccurate $z_s$.
This is not desirable because the attention module has to learn well-represented $z_s$ that can perform well on the source domain in order to capture the correct inter-pixel correlation of the source domain. The skip connection prevents this problem and enables the segmentation network to be trained good enough to perform segmentation correctly on the source domain.

\subsection{Domain adaptive training via SAM}
\label{da_SAM}
In this section, we explain the domain adaptive training of the segmentation network using the SAM that is pre-trained on the source domain. The overall training process is shown in Fig.~\ref{fig:Trainig} (b). Please note that the segmentation network in Fig.~\ref{fig:Trainig} (b) is a different network from the one in Fig.~\ref{fig:Trainig} (a). We train another separate segmentation network from scratch and this network is the one to be tested for the performance measurement.
Since this step is the actual process of domain adaptation, we train the network for both the source and the target doamin datasets. We input images of both domains into the network and get $z_s$ and $z_t$ respectively.
Then, $z_s$ and $z_t$ are used for two different loss functions. One is the commonly used segmentation loss previously introduced in Sec. \ref{preliminary}. More specifically, we upsample and apply softmax on $z_s$ and $z_t$ to get $p_s$ and $p_t$ to compute (\ref{eq:seg_src}) and (\ref{eq:seg_trg}). The other loss function is our proposed self-attention loss. $z_s$ and $z_t$ are given as inputs to the SAM that has already been pre-trained on the source domain. Inform that the pre-trained SAM is frozen and its weights are not updated during the domain adaptive training of the network. The SAM produces $z'_s$ and $z'_t$ which are the self-attended logits. We add $z_s$ and $z_t$ to them respectively and get $z''_s$ and $z''_t$. The proposed self-attention loss is defined as the $L1$ loss between $z$ and $z'$ or between $z$ and $z''$.
\vspace{-1mm}
\begin{equation}
    \mathcal{L}_{att}^S(x_s) = || z_s - z''_s ||^1_1 \quad \mathcal{L}_{att}^T(x_t) = || z_t - z''_t ||^1_1 \label{att_loss_1}
    \vspace{-1mm}
\end{equation}
\begin{equation}
    \mathcal{L}_{att}^S(x_s) = || z_s - z'_s ||^1_1 \quad \mathcal{L}_{att}^T(x_t) = || z_t - z'_t ||^1_1 \\ \label{att_loss_2}
    \vspace{-1mm}
\end{equation}
The self-attention loss is defined in two forms because we empirically observe that the self-attention loss shows different performances for different source datasets. The intention of the self-attention loss is to train the segmentation network to learn the self-attended output which contains information about inter-pixel correlations by forcing $z$ to mimic the output of the module, $z'$ or $z''$. When computing the self-attention loss, we detach the output of the module, $z'$ and $z''$, so the loss signal only flows through $z$ during the back-propagation. The overall objective functions are defined as follows:
\vspace{-1mm}
\begin{equation}
    \mathcal{L}= \mathcal{L}_{seg}^S(x_s) + \lambda\mathcal{L}_{att}^S(x_s) + \mathcal{L}_{seg}^T(x_t) + \lambda\mathcal{L}_{att}^T(x_t).
    \label{overall_1}
    \vspace{-1mm}
\end{equation}
\begin{equation}
    \mathcal{L}= \mathcal{L}_{seg}^S(x_s) + \mathcal{L}_{seg}^T(x_t) + \lambda\mathcal{L}_{att}^T(x_t).
    \label{overall_2}
    \vspace{-1mm}
\end{equation}
$\lambda$ is a hyper-parameter that balances between the segmentation loss and the self-attention loss. We empirically find that for GTA5\textrightarrow Cityscapes task, using (\ref{att_loss_1}) and applying self-attention loss for both domains, (\ref{overall_1}), performs better but for SYNTHIA\textrightarrow Cityscapes, using (\ref{att_loss_2}) and applying it only on the target domain, (\ref{overall_2}), shows better results. This will be further studied in the ablation study in Sec.~\ref{ablation}. In the inference time, prediction is based only on $z$, which is the output of the segmentation network without using the SAM.

\section{Experiments}
\label{experiment}

\subsection{Datasets and Training Details}
\textbf{Datasets.} We conduct UDA experiments using two different source domain datasets, GTA5 \cite{Richter_2016_ECCV} and SYNTHIA \cite{RosCVPR16} which are photo-realistic synthetic datasets collected from virtual games scenes. The target domain is the Cityscapes \cite{Cordts2016Cityscapes} which consists of real scenes under driving scenarios. The GTA5 contains 24,966 images and shares 19 classes with the Cityscapes while the SYNTHIA dataset contains 9,400 images and shares 16 classes in common with the Cityscapes. We additionally report the results of 13 common classes for SYNTHIA. The Cityscapes dataset contains 2,975 training images and 500 validation images. We conduct experiments on the validation set following the standard protocols of previous works.

\textbf{Training Details.} The experiments are mainly conducted using DeepLabV2 \cite{chen2017deeplab} with ResNet101 \cite{he2016deep} backbone. We also conduct experiment using FCN-8s \cite{long2015fully} with VGG16 backbone \cite{simonyan2014very} but its results and training details are in the supplementary. We initialize the backbone of the network with ImageNet pre-trained weights. We use the SGD optimizer with initial learning rate of $2.5 \times 10^{-4}$ and weight-decay of 0.0005. Learning rate is scheduled using `poly' learning rate policy with a power of 0.9. During the self-attention module training phase, both segmentation network and the SAM are jointly trained by the same optimizer. When domain adaptive training the segmentation network via the self-attention module, we use the pre-trained self-attention module that is saved at the last save point. The network is trained for 120,000 iterations with a batch size of 1. It is tested and saved at every 2,000 iterations. The hyper-parameter $\lambda$ is set to 0.1 empirically. We utilize the source images that are style-transferred into Cityscapes by \cite{CycleGAN2017} to further close the domain gap at image level.

\begin{table}[t]
		\begin{center}
		    \resizebox{0.9\linewidth}{!}{
			\begin{tabular}{l|c|c}
				\toprule
				Method & GTA5\textrightarrow CS & SYNTHIA\textrightarrow CS \\ 
				\midrule
				No Pseudo & 46.8 & 39.3 \\
				Pseudo-Only & 48.9 & 43.1 \\
				Ours (\ref{att_loss_1}) \& (\ref{overall_1}) & \textbf{50.7} & 43.5 \\
				Ours (\ref{att_loss_2}) \& (\ref{overall_1}) & 50.3 & 44.4 \\
				Ours Only on Target (\ref{overall_2}) & 50.3 (\ref{att_loss_1}) & \textbf{45.1} (\ref{att_loss_2}) \\
				No Conv & 42.7 & 44.0 \\
				No skip-connection & 49.7 & 44.4 \\
				\bottomrule
			\end{tabular}}
		\end{center}
		\vspace{-5mm}
		\caption{Results of Ablation Studies. The numbers are mIoU of 19 classes and 16 classes for GTA5\textrightarrow CS and SYNTHIA\textrightarrow CS respectively.}
		\label{table:ablation}
		\vspace{-4mm}
\end{table}

\begin{table}[t]
		\begin{center}
		    \resizebox{0.9\linewidth}{!}{
			\begin{tabular}{l|c|c|c|c}
				\toprule
				 & \multicolumn{2}{c|}{GTA5\textrightarrow CS} & \multicolumn{2}{c}{SYNTHIA\textrightarrow CS}  \\ 
				\midrule
				Gen & Pseudo-Only & Ours & Pseudo-Only & Ours \\ 
				\midrule
				Gen1 & 48.9 & 50.7 & 43.1 & 45.1 \\
				Gen2 & \textbf{49.7} & 51.2 & 45.2 & 47.9 \\
				Gen3 & 49.6 & 51.4 & 45.7 & 49.0 \\
				Gen4 & 49.1 & 51.7 & \textbf{45.9} & \textbf{49.2} \\
				Gen5 & 48.8 & \textbf{52.2} & 45.7 & 48.5 \\
				\bottomrule
			\end{tabular}}
		\end{center}
		\vspace{-5mm}
		\caption{Results of Iterative Training. The best results are in bold. mIoU 19 and mIoU 16 are used for GTA5\textrightarrow CS and SYNTHIA\textrightarrow CS respectively.}
		\vspace{-5mm}
		\label{table:iterative}
\end{table}

\begin{table*}[t]
		\setlength\tabcolsep{0.15em}
		\begin{center}
		    \resizebox{1.0\textwidth}{!}{
			\begin{tabular}{ @{} l|c|*{19}{c}|*{1}{c} @{} }
				\toprule
				Task & Method & {road} & {side.} & {build.} & \textbf{wall} & \textbf{fence} & \textbf{pole} & \textbf{light} & \textbf{sign} & {vege.} & \textbf{terrain} & {sky} & {person} & \textbf{rider} & {car} & \textbf{truck} & \textbf{bus} & \textbf{train} & \textbf{motor} & \textbf{bike} & {mIoU} \\
				\midrule
				{\multirow{2}{*}{GTA5\textrightarrow CS}} & Pseudo-Only & 90.6 & 43.7 & 84.6 & 37.6 & 24.8 & 33.8 & 39.4 & 36.9 & 84.6 & 38.8 & 84.3 & 59.7 & 26.0 & 85.3 & 39.5 & 50.0 & 0.0 & 24.6 & 45.1 & 48.9 \\
				& Ours & \textcolor{red}{91.0} & \textcolor{red}{46.9} & \textcolor{red}{85.7} & 37.6 & \textcolor{red}{25.5} & \textcolor{red}{34.2} & 39.4 & \textcolor{red}{37.7} & \textcolor{red}{85.3} & \textcolor{blue}{36.3} & \textcolor{red}{85.4} & \textcolor{red}{62.9} & \textcolor{red}{33.9} & \textcolor{red}{85.4} & \textcolor{red}{42.9} & \textcolor{red}{50.9} & 0.0 & \textcolor{red}{33.4} & \textcolor{red}{48.1} & \textcolor{red}{50.7} \\
				\midrule
				{\multirow{2}{*}{SYNTHIA \textrightarrow CS}} & Pseudo-Only & 72.7 & 29.2 & 80.0 & 13.8 & 0.8 & 31.3 & 24.5 & 28.4 & 80.3 & \textemdash & 82.4 & 50.9 & 24.7 & 74.0 & \textemdash & 24.4 & \textemdash & 21.7 & 49.9 & 43.1 \\
				& Ours & \textcolor{blue}{72.5} & \textcolor{blue}{28.6} & \textcolor{red}{82.2} & \textcolor{blue}{11.7} & \textcolor{blue}{0.7} & \textcolor{red}{33.8} & \textcolor{red}{29.5} & \textcolor{red}{28.9} & \textcolor{red}{81.7} & \textemdash & \textcolor{red}{85.0} & \textcolor{red}{56.1} & \textcolor{red}{27.3} & \textcolor{red}{77.7} & \textemdash & \textcolor{red}{29.9} & \textemdash & \textcolor{red}{24.5} & \textcolor{red}{50.1} & \textcolor{red}{45.1} \\
				\bottomrule
			\end{tabular}}
		\end{center}
		\vspace{-5mm}
		\caption{Effectiveness of `Ours' by classes. \textcolor{red}{Red} and \textcolor{blue}{Blue} refer to increase and decrease in performance. Bold faced are the rare classes.}
		\vspace{-5mm}
		\label{table:rare-classes}
\end{table*}

\subsection{Ablation Study}
\label{ablation}
Tab.\ref{table:ablation} shows our ablation studies. We test our proposed method in different settings and validate which setting works the best for each UDA task. `No Pseudo' does not employ any self-training via pseudo labels and only trains the network with the style-transferred source domain images. It shows the lowest performance for both UDA tasks. `Pseudo-Only' refers to applying self-training using the pseudo labels generated by the `No Pseudo' models. It is trained only by the two loss terms (\ref{eq:seg_src}) and (\ref{eq:seg_trg}). It clearly shows improved performance on both tasks. `Ours (\ref{att_loss_1})' and `Ours (\ref{att_loss_2})' are our proposed method introduced in Sec.\ref{da_SAM}. We utilize the same pseudo labels used in `Pseudo-Only' and additionally employ our proposed self-attention loss to show the efficacy of our method. The performances are significantly improved when our method is applied. However, there are some different tendency between the two UDA tasks. For GTA5\textrightarrow Cityscapes, (\ref{att_loss_1}) shows better performance than (\ref{att_loss_2}) while (\ref{att_loss_2}) results in higher mIoU than (\ref{att_loss_1}) on SYNTHIA\textrightarrow Cityscapes. Also, when we apply the self-attention loss only on the target domain (`Ours Only on Target'), employing (\ref{overall_2}), the performance on SYNTHIA improves but it rather decreases on GTA5. 
We conjecture this is due to the larger domain gap between SYNTHIA and Cityscapes than GTA5 and Cityscapes. (See supplementary for further discussion about it.)
Therefore, we utilize (\ref{att_loss_1}) and (\ref{overall_1}) on GTA5\textrightarrow Cityscapes while (\ref{att_loss_2}) and (\ref{overall_2}) on SYNTHIA\textrightarrow Cityscapes for the studies of `No Conv' and 'No Skip-connection'. Each refers to different architecture setting of SAM, `No Conv' trains the network using a SAM without the $1\times1$ convolutional layer, thus it does not need any training and the attention map is directly computed on $z$ without transformation. `No Skip-connection' trains the SAM without the skip-connection, hence the source segmentation loss is computed on $z'$ not on $z''$ during the SAM training phase. Without the skip-connection, the module fails to learn well-represented $z_s$ and it eventually under-performs during the domain adaptive training of the segmentation network. In both SAM settings, we could observe the performance degradation for both UDA tasks. This validates the necessity of the $1\times1$ convolutional layer and the skip-connection of the SAM.

\subsection{Iterative Training}
\label{iterative}
Self-training methods~\cite{li2019bidirectional, zhang2021prototypical, yang2020fda, zhang2019category} employ iterative training scheme which generates pseudo labels for the target domain from a trained segmentation network and retrains a new network with the generated pseudo labels. Tab.~\ref{table:iterative} shows the iterative training of `Pseudo-Only' and `Ours' on both UDA tasks. `Gen' stands for a generation which indicates the iteration of the iterative training. For example, Gen1 networks are trained with pseudo labels generated by the `No Pseudo' models from Tab.~\ref{table:ablation} and Gen2 networks are trained with pseudo labels generated by Gen1 networks.
On GTA5 task, the performance of `Pseudo-Only' shows the highest at Gen2 and rather decreases after as the generation goes on, while on SYNTHIA task, there is only a marginal improvement between the generations. On the other hand, when training with our proposed self-attention loss, the performance tends to improve better at every generation for both source domains. It implies that our proposed method helps the network not to produce noisy error-prone pseudo-labels that can lead to incorrect supervision signals and to generalize better on the target domain by transferring domain invariant inter-pixel correlations from the source domain. Note that the SAM is trained only \textit{once} on the source domain and the same SAM is reused at every generation for the proposed self-attention loss.

\subsection{Class-wise comparison between Pseudo-Only and Ours}
In Tab.~\ref{table:rare-classes} we compare our method with `Pseudo-Only' by classes to further investigate the effectiveness of our proposed method. As mentioned earlier, the same pseudo labels are utilized for both methods. `Ours' only additionally employs the proposed self-attention loss along with self-training. For both tasks, the mIoU is increased after applying our method. Our method particularly contributes to the performance gain on rare classes which scarcely exist in the dataset. Wall, fence, pole, light, sign, terrain, rider, truck, bus, train, motor and bike are known rare classes. On GTA5\textrightarrow Cityscapes, IoU of rider, truck, motor, and bike are significantly improved, besides, other rare classes such as fence, pole, sign and bus also show increased IoU. Similarly on SYNTHIA\textrightarrow Cityscapes, `Ours' shows better performance than `Pseudo-Only' on pole, light, sign, rider, bus, motor and bike. Other than rare-classes, our method also leads to significant improvement of majors classes such as sky, person and car on both tasks.

\begin{table*}[t]
		\setlength\tabcolsep{0.15em}
		\begin{center}
		    \resizebox{0.9\textwidth}{!}{
			\begin{tabular}{ l|*{19}{c}|c }
				\toprule
				Method & {road} & {side.} & {build.} & {wall} & {fence} & {pole} & {light} & {sign} & {vege.} & {terrain} & {sky} & {person} & {rider} & {car} & {truck} & {bus} & {train} & {motor} & {bike} & {mIoU} \\
				\midrule
				PLCA \cite{kang2020pixel} & 84.0 & 30.4 & 82.4 & 35.3 & 24.8 & 32.2 & 36.8 & 24.5 & 85.5 & 37.2 & 78.6 & 66.9 & 32.8 & 85.5 & 40.4 & 48.0 & 8.8 & 29.8 & 41.8 & 47.7 \\
				
				BDL \cite{li2019bidirectional} & 91.0 & 44.7 & 84.2 & 34.6 & 27.6 & 30.2 & 36.0 & 36.0 & 85.0 & 43.6 & 83.0 & 58.6 & 31.6 & 83.3 & 35.3 & 49.7 & 3.3 & 28.8 & 35.6 & 48.5 \\
				
				CrCDA \cite{huang2020contextual} & 92.4 & 55.3 & 82.3 & 31.2 & 29.1 & 32.5 & 33.2 & 35.6 & 83.5 & 34.8 & 84.2 & 58.9 & 32.2 & 84.7 & 40.6 & 46.1 & 2.1 & 31.1 & 32.7 & 48.6 \\
				
				SIM \cite{wang2020differential} & 90.6 & 44.7 & 84.8 & 34.3 & 28.7 & 31.6 & 35.0 & 37.6 & 84.7 & 43.3 & 85.3 & 57.0 & 31.5 & 83.8 & 42.6 & 48.5 & 1.9 & 30.4 & 39.0 & 49.2 \\
				
				CADA~\cite{yang2021context} & 91.3 & 46.0 & 84.5 & 34.4 & 29.7 & 32.6 & 35.8 & 36.4 & 84.5 & 43.2 & 83.0 & 60.0 & 32.2 & 83.2 & 35.0 & 46.7 & 0.0 & 33.7 & 42.2 & 49.2 \\
				
                Label-driven\cite{yang2020label} &
				90.8 & 41.4 & 84.7 & 35.1 & 27.5 & 31.2 & 38.0 & 32.8&85.6&42.1&84.9&59.6&34.4&85.0&42.8&52.7&3.4&30.9&38.1&49.5 \\
				
				MaxCos~\cite{chung2021maximizing} & 92.6 & 54.0 & 85.4 & 35.0 & 26.0 & 32.4 & 41.2 & 29.7 & 85.1 & 40.9 & 85.4 & 62.6 & 34.7 & 85.7 & 35.6 & 50.8 & 2.4 & 31.0 & 34.0 & 49.7 \\
                
				Kim et al. \cite{kim2020learning} & 92.9 & 55.0 & 85.3 & 34.2 & 31.1 & 34.9 & 40.7 & 34.0 & 85.2 & 40.1 & 87.1 & 61.0 & 31.1 & 82.5 & 32.3 & 42.9 & 0.3 & 36.4 & 46.1 & 50.2 \\
				
				CAG\_UDA~\cite{zhang2019category}  & 90.4& 51.6& 83.8& 34.2& 27.8& 38.4& 25.3& 48.4& 85.4& 38.2& 78.1& 58.6& 34.6& 84.7& 21.9& 42.7& \bf41.1& 29.3& 37.2& 50.2 \\
				
				Seg-Uncertainty~\cite{zheng2020rectifying} &90.4& 31.2& 85.1& 36.9& 25.6& 37.5& 48.8& 48.5& 85.3& 34.8& 81.1& 64.4& 36.8& 86.3& 34.9& 52.2& 1.7& 29.0& 44.6& 50.3 \\
				
				FDA-MBT \cite{yang2020fda} & 92.5 & 53.3 & 82.4 & 26.5 & 27.6 & 36.4 & 40.6 & 38.9 & 82.3 & 39.8 & 78.0 & 62.6 & 34.4 & 84.9 & 34.1 & 53.1 & 16.9 & 27.7 & 46.4 & 50.5 \\
				
				TPLD~\cite{shin2020two} & \bf94.2 & \bf60.5 & 82.8 & 36.6 & 16.6 & 39.3 & 29.0 & 25.5 & 85.6 & 44.9 & 84.4 & 60.6 & 27.4 & 84.1 & 37.0 & 47.0 & 31.2 & 36.1 & 50.3 & 51.2 \\
				
				MetaCorrection~\cite{guo2021meatcorrection} & 92.8 & 58.1 & 86.2 & 39.7 & 33.1 & 36.3 & 42.0 & 38.6 & 85.5 & 37.8 & 87.6 & 62.8 & 31.7 & 84.8 & 35.7 & 50.3 & 2.0 & 36.8 & 48.0 & 52.1 \\
				
				\midrule
				Ours & 90.8 & 47.2 & \bf86.8 & 41.5 & 29.4 & 35.7 & 42.4 & 37.4 & 86.0 & 42.1 & \bf88.3 & 63.7 & 35.6 & 85.1 & 43.8 & 54.6 & 0.0 & 33.6 & 47.8 & 52.2 \\
				
				\midrule
				ProDA \cite{zhang2021prototypical} & 87.8 & 56.0 & 79.7 & \bf46.3 & 44.8 & 45.6 & \bf53.5 & 53.5 & 88.6 & 45.2 & 82.1 & \bf70.7 & \bf39.2 & 88.8 & 45.5 & 59.4 & 1.0 & \bf48.9 & \bf56.4 & 57.5 \\
				
				ProDA+Ours & 89.3 & 56.4 & 81.1 & 46.1 & \bf46.2 & \bf47.0 & 52.9 & \bf54.1 & \bf88.7 & \bf48.3 & 82.5 & 70.4 & 39.1 & \bf89.3 & \bf51.8 & \bf61.1 & 0.1 & 44.4 & 55.0 & \bf58.1 \\
				\bottomrule
			\end{tabular}}
		\end{center}
		\vspace{-5mm}
		\caption{Comparison results with other methods on GTA5\textrightarrow Cityscapes. The numbers in bold are the best score for each column.}
		\vspace{-2mm}
		\label{table:gta5}
\end{table*}

\begin{table*}[t]
		\setlength\tabcolsep{0.15em}
		\begin{center}
		    \resizebox{0.9\textwidth}{!}{
			\begin{tabular}{ l|*{16}{c}|c|c }
				\toprule
				Method & {road} & {side.} & {build.} & {wall} & {fence} & {pole} & {light} & {sign} & {vege.} & {sky} & {person} & {rider} & {car} & {bus} & {motor} & {bike} & {mIoU*} & {mIoU} \\
				\midrule
				Kim et al. \cite{kim2020learning} & \bf92.6 & \bf53.2 & 79.2 & \textemdash & \textemdash & \textemdash & 1.6 & 7.5 & 78.6 & 84.4 & 52.6 & 20.0 & 82.1 & 34.8 & 14.6 & 39.4 & 49.3 & \textemdash \\
				
				CrCDA \cite{huang2020contextual} & 86.2 & 44.9 & 79.5 & 8.3 & 0.7 & 27.8 & 9.4 & 11.8 & 78.6 & \bf86.5 & 57.2 & 26.1 & 76.8 & 39.9 & 21.5 & 32.1 & 50.0 & 42.9 \\
				
				BDL \cite{li2019bidirectional} &
				86.0 & 46.7 & 80.3 & \textemdash & \textemdash & \textemdash & 14.1 & 11.6 & 79.2 & 81.3 & 54.1 & 27.9 & 73.7 & 42.2 & 25.7 & 45.3 & 51.4 & \textemdash \\
				
				CAG\_UDA~\cite{zhang2019category}  &84.7& 40.8& 81.7& 7.8& 0.0& 35.1& 13.3& 22.7& 84.5& 77.6& 64.2& 27.8& 80.9& 19.7& 22.7& 48.3 & 51.5 & 44.5 \\
				
				SIM \cite{wang2020differential} &
				83.0 & 44.0 & 80.3 & \textemdash & \textemdash &  \textemdash & 17.1 & 15.8 & 80.5 & 81.8 & 59.9 & \bf33.1 & 70.2 & 37.3 & 28.5 & 45.8 & 52.1 & \textemdash \\
				
				CADA~\cite{yang2021context} & 82.5 & 42.2 & 81.3 & \textemdash & \textemdash &  \textemdash & 18.3 & 15.9 & 80.6 & 83.5 & 61.4 & 33.2 & 72.9 & 39.3 & 26.6 & 43.9 & 52.4 & \textemdash \\
				
				FDA-MBT \cite{yang2020fda} & 79.3 & 35.0 & 73.2 & \textemdash & \textemdash &  \textemdash & 19.9 & 24.0 & 61.7 & 82.6 & 61.4 & 31.1 &  83.9 & 40.8 & 38.4 & 51.1 & 52.5 & \textemdash \\
				
				MetaCorrection~\cite{guo2021meatcorrection} & \bf92.6 & 52.7 & 81.3 & 8.9 & \bf2.4 & 28.1 & 13.0 & 7.3 & 83.5 & 85.0 & 60.1 & 19.7 & 84.8 & 37.2 & 21.5 & 43.9 & 52.5 & 45.1 \\
				
				Label-driven\cite{yang2020label} & 85.1 & 44.5 & 81.0 & \textemdash & \textemdash & \textemdash & 16.4 & 15.2 & 80.1 & 84.8 & 59.4 & 31.9 & 73.2 & 41.0 & 32.6 & 44.7 & 53.1 & \textemdash \\
				
				MaxCos~\cite{chung2021maximizing} & 88.3 & 47.3 & 80.1 & \textemdash & \textemdash & \textemdash& 21.6 & 20.2 & 79.6 & 82.1 & 59.0 & 28.2 & 82.0 & 39.2 & 17.3 & 46.7 & 53.2 & \textemdash \\
				
				PLCA\cite{kang2020pixel} & 82.6 & 29.0 & 81.0 & 11.2 & 0.2 & 33.6 & 24.9 & 18.3 & 82.8 & 82.3 & 62.1 & 26.5 & 85.6 & 48.9 & 26.8 & 52.2 & 54.0 & 46.8 \\
				
                TPLD~\cite{shin2020two} & 80.9 & 44.3 & 82.2 & 19.9 & 0.3 & 40.6 & 20.5 & 30.1 & 77.2 & 80.9 & 60.6 & 25.5 & 84.8 & 41.1 & 24.7 & 43.7 & 53.5 & 47.3  \\
				
                Seg-Uncertainty~\cite{zheng2020rectifying} & 87.6& 41.9& 83.1& 14.7& 1.7& 36.2& 31.3& 19.9& 81.6& 80.6& 63.0& 21.8& 86.2& 40.7& 23.6 & 53.1 & 54.9 & 47.9 \\
                
				\midrule
				Ours & 77.5 & 32.3 & 82.6 & 25.5 & 1.9 & 34.6 & 33.6 & 32.4 & 81.7 & 85.1 & 63.8 & 31.8 &  82.3 & 35.2 & 31.9 & \bf54.6 & 55.7 & 49.2 \\
				
				\midrule
				ProDA~\cite{zhang2021prototypical} & 87.8 & 45.7 & 84.6 & 37.1 & 0.6 & 44.0 & \bf54.6 & 37.0 & 88.1 & 84.4 & \bf74.2 & 24.3 & 88.2 & 51.1 & 40.5 & 45.6 & 62.0 & 55.5  \\
				
				ProDA~\cite{zhang2021prototypical} + Ours & 87.3 & 45.4 & \bf85.3 & \bf37.3 & 0.3 & \bf46.2 & 54.2 & \bf39.0 & \bf88.3 & 84.6 & 73.7 & 22.9 & \bf88.5 & \bf59.8 & \bf41.8 & 44.1 & \bf62.7 & \bf56.2 \\
				\bottomrule
			\end{tabular}}
		\end{center}
		\vspace{-5mm}
		\caption{Comparison results with other methods on SYNTHIA\textrightarrow Cityscapes. The numbers in bold are the best score for each column. mIoU* and mIoU denote mIoU of 13 classes and 16 classes respectively.}
		\vspace{-5mm}
		\label{table:synthia}
\end{table*}

\subsection{Comparisons with state-of-the-art methods}
Tab.~\ref{table:gta5} and Tab.~\ref{table:synthia} show the comparison experiments with other state-of-the-art methods on both datasets. Our method outperforms other methods except for ProDA~\cite{zhang2021prototypical} which achieved a major breakthrough in the UDA for semantic segmentation. Since our proposed method can be easily applied to other methods by simply adding our self-attention loss term to the overall objective function, we combine our method with ProDA by employing our self-attention loss during the stage3 training phase of ProDA. After applying our method on ProDA, we could observe an additional performance gain for both UDA tasks. On GTA5\textrightarrow Cityscapes, 12 out of 19 classes show increased IoU, especially, the performance on rare classes such as fence, pole, terrain, truck and bus are highly improved. Similar results can be found on SYNTHIA\textrightarrow Cityscapes as well, mIoU is improved along with IoU of some rare-classes. Also, it is noteworthy that when excluding ProDA, there is a more than 1\% gap between `Ours' (49.2\%) and the second-best method (47.9\%) on SYNTHIA task. We empirically find that $\lambda$ = 0.005 works the best when combining our method with ProDA. On both UDA benchmarks, our method achieves comparable performance to existing approaches. Moreover, we show that our method can be easily applied to other method and validate its effectiveness by additionally employing it on leading SOTA method and setting a new SOTA performance for both UDA tasks.

\begin{figure*}[t]
    \centering
    \includegraphics[width = 1.0\linewidth]{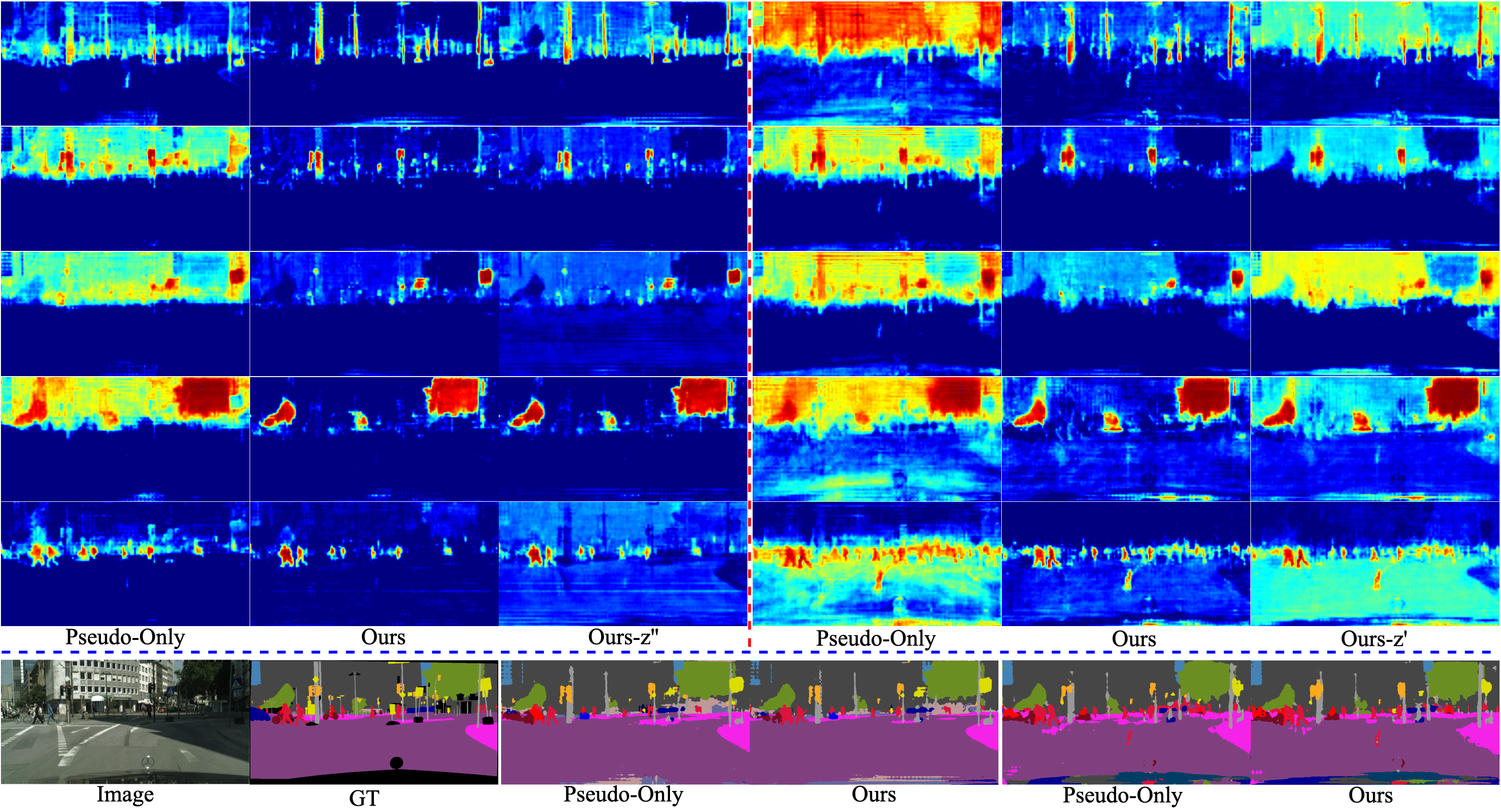}
    \vspace{-6mm}
    \caption{The attention and prediction visualization. Images above the \textcolor{blue}{blue} dashed line are attention visualization and ones under the line are prediction visualization. In the attention visualization, the images on the left side of the \textcolor{red}{red} dashed line are the results of GTA5\textrightarrow Cityscapes while ones on the right are SYNTHIA\textrightarrow Cityscapes. Each row of attention visualization refers to 'pole', 'light', 'sign', 'vege.' and 'person' from top to bottom. One image is used to fairly show the difference between different classes and methods.}
    \vspace{-4mm}
    \label{fig:qualitative_analysis}
\end{figure*}

\subsection{Qualitative Analysis}
\label{qualitative}
Fig.~\ref{fig:qualitative_analysis} shows attention and prediction visualization of `Pseudo-Only' and `Ours'. `Pseudo-Only' and `Ours' are networks tested in Tab.~\ref{table:rare-classes}. The attention is visualized as the similarity between the pixel with the highest confidence of the class we want to measure and all the other pixels in the image. The most confident pixel is marked as a blue dot in the visualization. The redder color means higher similarity while the bluer color refers to lower similarity. If the network is trained well, the most confident pixel should have high similarity only with the pixels belonging to the same class. We additionally visualize the results after applying the SAM using $z''$ and $z'$ (`Ours-$z''$' \& `Ours-$z'$') for GTA5 and SYNTHIA respectively. 

When using only the self-training (`Pseudo-Only'), the most confident pixel has high similarity even with irrelevant pixels of different classes, while 'Ours' is more attended well with the relevant pixels corresponding to the same class. The results of SYNTHIA are more noisy due to the larger domain gap. `Ours-$z''$' and `Ours-$z'$' look a little more noisy than `Ours' but clearer and more distinct than `Pseudo-Only'. They give us hint about what knowledge is being transferred via our proposed self-attention loss. We conjecture that their little noisiness provides extra information about inter-pixel correlations to the network. In the prediction visualization, the 3rd and the 4th columns correspond to GTA5 task while the 5th and the 6th columns refer to SYNTHIA task. When comparing only the prediction results, it is hard to see the difference between the two methods, even though `Ours' look a little more accurate. However, when we visualize the attention by classes and compare them as above, we can clearly observe a distinct difference between the two. It implies that even though their prediction visualization might seem similar, their predicted logits have very different distribution which leads to better performance for `Ours'. These results are not cherry-picked and more results using different images can be found in the supplementary.

\begin{figure}[t]
    \centering
    \includegraphics[width = 1.0\linewidth]{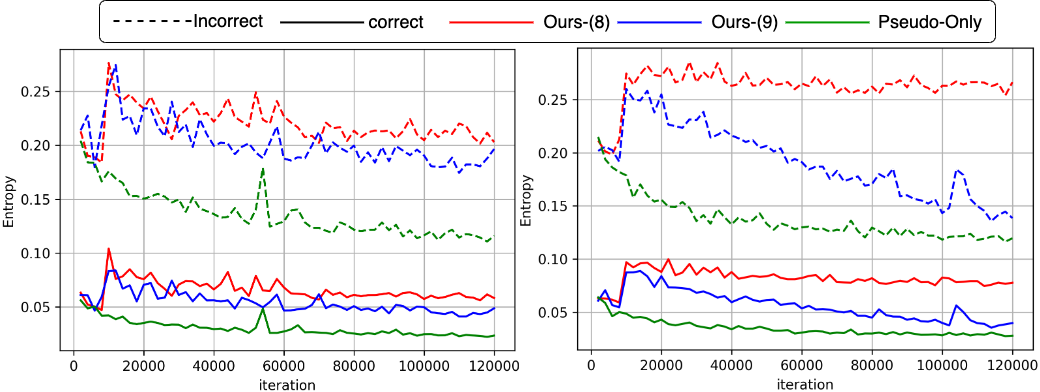}
    \vspace{-5mm}
    \caption{Entropy analysis of GTA5\textrightarrow Cityscapses (Left) and SYNTHIA\textrightarrow Cityscapses (right).}
    \vspace{-5mm}
    \label{fig:quantitative_analysis}
\end{figure}

\subsection{Quantitative Analysis}
Fig.~\ref{fig:quantitative_analysis} shows entropy analysis of three different networks on the two UDA tasks. Every 2,000 iteration, we measure the entropy of all the 500 validation images of Cityscapes. For each image, we divide the pixels into two groups, correctly and incorrectly classified, and compute the entropy for each group as follows:
\vspace{-2mm}
\begin{equation}
    Ent(x_t) = \frac{1}{\# \text{pixel}}(-\sum_{i=1}^{\# \text{pixel}} \sum_{c=1}^C p_t^{(i,c)}\log (p_t^{(i,c)})) \\
    \vspace{-2mm}
\end{equation}
We plot the average entropy of the 500 images for each pixel group separately in the figure. The solid line is the entropy of correct pixels and the dashed line refers to the incorrect pixels. Each red, blue and green line indicates `Ours-(\ref{att_loss_1})', `Ours-(\ref{att_loss_2})' and `Pseudo-Only' networks respectively. For both benchmarks and all three networks, incorrect pixels have higher entropy than correct pixels. It is an interesting observation that `Pseudo-Only' has the lowest entropy while `Ours-(\ref{att_loss_1})' and `Ours-(\ref{att_loss_2})' show higher entropy than `Pseudo-Only' throughout the training. Having higher entropy means that the class distribution of a pixel is not biased towards a certain class (a highly peaky distribution) but rather has probability for some of other classes as well. We conjecture this is due to our self-attention loss providing extra knowledge about inter-pixel correlations which prevents the network from being highly biased and over-confident.

\vspace{-2mm}
\section{Conclusion}
In this paper, we propose a method of transferring domain invariant inter-pixel correlations from the source domain to the target domain by utilizing a self-attention module that is pre-trained on the source domain. Our method is very simple and does not require complex and heuristic algorithm (only one hyper-parameter, $\lambda$, to tune) yet very effective. Our method supports self-training by overcoming the defects of noisy pseudo labels with extra knowledge provided by the SAM. Also, we show that our method can be easily used in addition to other methods to further enhance the performance. The proposed method sets a new SOTA scores on both UDA benchmarks when combined with the recent SOTA method~\cite{zhang2021prototypical}.

\appendix
\counterwithin{figure}{section}
\counterwithin{table}{section}

\section{Experiments on FCN-8s with VGG16 backbone}

\subsection{Training details}
We also conduct experiments using a FCN-8s with VGG16 backbone which is another segmentation network that is widely used in unsupervised domain adaptation for semantic segmentation. We train FCN-8s with VGG16 backbone by ADAM optimizer with an initial learning rate of $1 \times 10^{-5}$ and the momentums of 0.9 and 0.99. The learning rate is decayed by `step' learning rate policy with a step size of 50,000 and a decay rate of 0.1. The hyper-parameter $\lambda$ is set to 0.001 empirically. Other details are the same as the DeepLabV2 with ResNet101 backbone. Different from DeepLabV2, FCN-8s uses transposed convolution instead of bilinear interpolation for upsampling $z$.

\begin{table}[t]
		\begin{center}
		    \resizebox{0.95\linewidth}{!}{
			\begin{tabular}{l|c|c}
				\toprule
				Method & GTA5\textrightarrow CS & SYNTHIA\textrightarrow CS \\ 
				\midrule
				No Pseudo & 40.6 & 36.3 \\
				Pseudo-Only & 42.8 & 39.6 \\
				Ours (\ref{att_loss_1}) \& (\ref{overall_1}) & \bf43.1 & \bf40.0 \\
				Ours (\ref{att_loss_2}) \& (\ref{overall_1}) & 42.8 & 39.7 \\
				Ours only on Target (\ref{overall_2}) &  42.6(\ref{att_loss_1}) &  39.7(\ref{att_loss_1}) \\
				\bottomrule
			\end{tabular}}
		\end{center}
		\vspace{-5mm}
		\caption{Results of Ablation Studies. The numbers are mIoU of 19 classes and 16 classes for GTA5\textrightarrow CS and SYNTHIA\textrightarrow CS respectively.}
		\vspace{-5mm}
		\label{table:vgg_ablation}
\end{table}

\subsection{Ablation Study}
Tab. \ref{table:vgg_ablation} shows our ablation study on FCN-8s with VGG16 backbone. As shown in the table, applying our self-attention loss improves the performance for both domains, but its performance gain is somewhat lower than that of DeepLabV2. Moreover, in contrast to DeepLabV2, using (\ref{att_loss_1}) instead of (\ref{att_loss_2}) and applying our self-attention loss on both domains rather than only on the target domain achieve better performance when using FCN-8s. We conjecture this difference comes from the different architecture of FCN-8s which uses transposed convolution for upsampling $z$ instead of bilinear interpolation used in DeepLabV2. Since our proposed self-attention loss is computed on $z$ and not on $U(z)$, it can not train the transposed convolutional layer of FCN-8s. This could be the possible reason why ours does not show as much performance gain as it shows in the DeepLabV2 experiment.

\begin{table}[t]
		\begin{center}
		    \resizebox{0.95\linewidth}{!}{
			\begin{tabular}{l|c|c|c|c}
				\toprule
				 & \multicolumn{2}{c|}{GTA5\textrightarrow CS} & \multicolumn{2}{c}{SYNTHIA\textrightarrow CS}  \\ 
				\midrule
				Gen & Pseudo-Only & Ours & Pseudo-Only & Ours \\ 
				\midrule
				Gen1 & 42.8 & 43.1 & 39.6 & 40.0 \\
				Gen2 & 43.0 & 43.9 & 40.7 & 41.5 \\
				Gen3 & 43.3 & 44.2 & 41.0 & 42.0 \\
				Gen4 & 44.1 & 44.7 & 42.1 & 42.2 \\
				Gen5 & 44.2 & 45.7 & \bf42.4 & 42.8 \\
				Gen6 & \bf44.3 & \bf45.7 & 42.3 & \bf43.1 \\
				\bottomrule
			\end{tabular}}
		\end{center}
		\vspace{-5mm}
		\caption{Results of Iterative Training. The best results are in bold. mIoU 19 and mIoU 16 are used for GTA5\textrightarrow CS and SYNTHIA\textrightarrow CS respectively.}
		\vspace{-5mm}
		\label{table:vgg_iterative}
\end{table}

\subsection{Iterative Training}
We also conduct iterative training analogous to DeepLabV2. As it can be seen in the Tab. \ref{table:vgg_iterative}, Both `Pseudo-Only' and `Ours' show improved performance as the generation goes on. However, `Ours' shows better performance improvement than `Pseudo-Only' between the generations. The performance of `Pseudo-Only' gets saturated around a certain generation while `Ours' keeps showing noticeable performance gain even in the later generation.

\begin{table*}[t]
		\setlength\tabcolsep{0.15em}
		\begin{center}
		    \resizebox{0.95\textwidth}{!}{
			\begin{tabular}{ l|*{19}{c}|c }
			    \toprule
				Method & {road} & {side.} & {build.} & {wall} & {fence} & {pole} & {light} & {sign} & {vege.} & {terrain} & {sky} & {person} & {rider} & {car} & {truck} & {bus} & {train} & {motor} & {bike} & {mIoU} \\
				
				\midrule
				CrDoCo \cite{chen2019crdoco} & 89.1 & 33.2 & 80.1 & 26.9 & 25.0 & 18.3 & 23.4 & 12.8 & 77.0 & 29.1 & 72.4 & 55.1 & 20.2 & 79.9 & 22.3 & 19.5 & 1.0 & 20.1 & 18.7 & 38.1 \\
				
				CrCDA \cite{huang2020contextual} & 86.8 & 37.5 & 80.4 & 30.7 & 18.1 & 26.8 & 25.3 & 15.1 & 81.5 & 30.9 & 72.1 & 52.8 & 19.0 & 82.1 & 25.4 & 29.2 & 10.1 & 15.8 & 3.7 & 39.1 \\
				
				BDL \cite{li2019bidirectional} & 89.2 & 40.9 & 81.2 & 29.1 & 19.2 & 14.2 & 29.0 & 19.6 & 83.7 & 35.9 & 80.7 & 54.7 & 23.3 & 82.7 & 25.8 & 28.0 & 2.3 & 25.7 & 19.9 & 41.3 \\  
				
				FDA-MBT \cite{yang2020fda} & 86.1 & 35.1 & 80.6 & 30.8 & 20.4 & 27.5 & 30.0 & 26.0 & 82.1 & 30.3 & 73.6 & 52.5 & 21.7 & 81.7 & 24.0 & 30.5 & \bf29.9 & 14.6 & 24.0 & 42.2 \\
				
				Kim et al. \cite{kim2020learning} & \bf92.5 & \bf54.5 & \bf83.9 & \bf34.5 & \bf25.5 & \bf31.0 & 30.4 & 18.0 & 84.1 & 39.6 & \bf83.9 & 53.6 & 19.3 & 81.7 & 21.1 & 13.6 & 17.7 & 12.3 & 6.5 & 42.3 \\
				
				SIM \cite{wang2020differential} & 88.1 & 35.8 & 83.1 & 25.8 & 23.9 & 29.2 & 28.8 & 28.6 & 83.0 & 36.7 & 82.3 & 53.7 & 22.8 & 82.3 & 26.4 & \bf38.6 & 0.0 & 19.6 & 17.1 & 42.4 \\
				
				Label-driven\cite{yang2020label} & 90.1 & 41.2 & 82.2 & 30.3 & 21.3 & 18.3 & 33.5 & 23.0 & 84.1 & 37.5 & 81.4 & 54.2 & 24.3 & 83.0 & 27.6 & 32.0 & 8.1 & \bf29.7 & 26.9 & 43.6 \\
				
				MaxCos~\cite{chung2021maximizing} & 90.3 & 42.6 & 82.2 & 29.7 & 22.2 & 18.5 & 32.8 & 26.8 & 84.3 & 37.1 & 80.2 & 55.2 & 26.4 & 83.0 & \bf30.3 & 35.1 & 7.0 & 29.6 & 28.9 & 44.3 \\
				
				CADA~\cite{yang2021context} & 90.1 & 46.7 & 82.7 & 34.2 & 25.3 & 21.3 & 33.0 & 22.0 & \bf84.4 & 41.4 & 78.9 & 55.5 & 25.8 & \bf83.1 & 24.9 & 31.4 & 20.6 & 25.2 & 27.8 & 44.9 \\
				
				\midrule
				Ours & 87.4 & 40.8 & 81.8 & 31.7 & 19.3 & 26.3 & \bf36.3 & \bf34.1 & 83.9 & \bf43.2 & 79.9 & \bf56.1 & \bf27.0 & 81.8 & 26.4 & 38.3 & 4.1 & 29.4 & \bf39.9 & \bf45.7 \\   
				
				\bottomrule
			\end{tabular}}
		\end{center}
		\vspace{-5mm}
		\caption{Comparison results with other methods on GTA5\textrightarrow Cityscapes. The numbers in bold are the best score for each column.}
		\vspace{-3mm}
		\label{table:vgg_gta5}
\end{table*}

\begin{table*}[t]
		\setlength\tabcolsep{0.15em}
		\begin{center}
		    \resizebox{0.95\textwidth}{!}{
			\begin{tabular}{ l|*{16}{c}|c|c }
				\toprule
				Method & {road} & {side.} & {build.} & {wall} & {fence} & {pole} & {light} & {sign} & {vege.} & {sky} & {person} & {rider} & {car} & {bus} & {motor} & {bike} & {mIoU*} & {mIoU} \\
				
				\midrule
				CrCDA \cite{huang2020contextual} & 74.5 & 30.5 & 78.6 & 6.6 & 0.7 & 21.2 & 2.3 & 8.4 & 77.4 & 79.1 & 45.9 & 16.5 & 73.1 & 24.1 & 9.6 & 14.2 & 41.1 & 35.2 \\
				
				ROAD-Net \cite{chen2018road} & 77.7 & 30.0 & 77.5 & 9.6 & 0.3 & 25.8 & 10.3 & 15.6 & 77.6 & 79.8 & 44.5 & 16.6 & 67.8 & 14.5 & 7.0 & 23.8 & 41.7 & 36.2 \\
				
				GIO-Ada \cite{chen2019learning} & 78.3 & 29.2 & 76.9 & \bf11.4 & 0.3 & 26.5 & 10.8 & 17.2 & \bf81.7 & \bf81.9 & 45.8 & 15.4 & 68.0 & 15.9 & 7.5 & 30.4 & 43.0 & 37.3 \\
				
				Kim et al.~\cite{kim2020learning} & \bf89.8 & \bf48.6 & 78.9 & \textemdash & \textemdash & \textemdash & 0.0 & 4.7 & 80.6 & 81.7 & 36.2 & 13.0 & 74.4 & 22.5 & 6.5 & 32.8 & 43.8 & \textemdash \\
				
				CrDoCo \cite{chen2019crdoco} & 84.9 & 32.8 & \bf80.1 & 4.3 & 0.4 & \bf29.4 & 14.2 & 21.0 & 79.2 & 78.3 & 50.2 & 15.9 & 69.8 & 23.4 & 11.0 & 15.6 & 44.3 & 38.2 \\
				
				BDL \cite{li2019bidirectional} & 72.0 & 30.3 & 74.5 & 0.1 & 0.3 & 24.6 & 10.2 & 25.2 & 80.5 & 80.0 & 54.7 & 23.2 & 72.7 & 24.0 & 7.5 & 44.9 & 46.1 & 39.0 \\
				
				FDA-MBT \cite{yang2020fda} & 84.2 & 35.1 & 78.0 & 6.1 & 0.44 & 27.0 & 8.5 & 
				22.1 & 77.2 & 79.6 & 55.5 & 19.9 & 74.8 & 24.9 & \bf14.3 & 40.7 & 47.3 & 40.5 \\
				
				CADA~\cite{yang2021context} & 73.0 & 31.1 & 77.1 & 0.2 & 0.5 & 27.0 & 11.3 &  27.4 & 81.2 & 81.0 & \bf59.0 & \bf25.6 & 75.0 & 26.3 & 10.1 & 47.4 & 48.1 & 40.8 \\
				
				Label-driven\cite{yang2020label} & 73.7 & 29.6 & 77.6 & 1.0 & 0.4 & 26.0 & 14.7 & 26.6 & 80.6 & 81.8 & 57.2 & 24.5 & 76.1 & 27.6 & 13.6 & 46.6 & 48.5 & 41.1 \\
				
				MaxCos~\cite{chung2021maximizing} & 73.6 & 30.6 & 77.5 & 0.8 & 0.4 & 26.7 & 14.1 & \bf29.3 & 80.9 & 80.6 & 57.9 & 24.7 & \bf76.5 & 27.2 & 10.8 & \bf47.8 & 48.6 & 41.2 \\
				
				\midrule
				Ours & 85.6 & 43.7 & 77.9 & 7.0 & \bf0.8 & 26.3 & \bf21.4 & 25.4 & 80.8 & 80.5 & 58.6 & 21.2 & 74.7 & \bf29.1 & 12.5 & 44.3 & \bf50.5 & \bf43.1 \\
				
				\bottomrule
			\end{tabular}}
		\end{center}
		\vspace{-5mm}
		\caption{Comparison results with other methods on SYNTHIA\textrightarrow Cityscapes. The numbers in bold are the best score for each column. mIoU* and mIoU denote mIoU of 13 classes and 16 classes respectively.}
		\vspace{-5mm}
		\label{table:vgg_synthia}
\end{table*}

\subsection{Comparison with other methods}
Tab.~\ref{table:vgg_gta5} and Tab.~\ref{table:vgg_synthia} show performance comparison with other methods using FCN-8s with VGG16 backbone on GTA5\textrightarrow Cityscapes and SYNTHIA\textrightarrow Cityscapes tasks respectively. Some methods are not included in the table because they only conduct experiments using the DeepLabV2. Our method achieves the highest performance compared to other state-of-the-art methods.

\begin{table}[t]
		\begin{center}
		    \resizebox{0.8\linewidth}{!}{
			\begin{tabular}{l|c|l|c}
				\toprule
				\multicolumn{2}{c|}{Loss function} & \multicolumn{2}{c}{Layer}  \\ 
				\midrule
				$L1$ & 50.7 & Prediction & 50.7 \\
				KL-Div & 49.2 & Feature map & 50.4 \\
				Cosine & 48.5 & Both & 50.2 \\
				\midrule
				\multicolumn{2}{c|}{Pseudo-Only} & \multicolumn{2}{c}{48.9} \\
				\bottomrule
			\end{tabular}}
		\end{center}
		\vspace{-5mm}
		\caption{Experimental results of using different loss functions and computing the loss at different layers of the segmentation network.}
		\vspace{-5mm}
		\label{table:loss_layer}
\end{table}

\section{Loss function and Layer Analysis}
We conduct experiment of computing the proposed self-attention loss utilizing different loss functions and at different layers of the segmentation network. The experiment is done on GTA5\textrightarrow Cityscapes task using the DeepLabV2 with ResNet101 backbone. `$L1$', `KL-Div' and `Cosine' refer to $L1$ loss, Kullback-Leibler divergence loss and cosine similarity loss respectively. `KL-Div' and `Cosine' are computed between each logit (row) of $z$ and $z''$. The cosine similarity is calculated between $z_i \in \mathbb{R}^{C}$ and $z''_i \in \mathbb{R}^{C}$ which are the logits of $z$ and $z''$ respectively. It trains the segmentation network by maximizing the computed cosine similarity or minimizing the negative cosine similarity equivalently. `KL-Div' loss function tries to minimize the KL-divergence between $z_i$ and $z''_i$. We first apply softmax on $z_i$ and $z''_i$ and then calculate the KL-divergence between them, $D_{KL}( \sigma(z''_i) || \sigma(z_i) )$, here $\sigma$ refers to the softmax. We train the segmentation network with three different loss functions independently and compare the results. As shown in the Tab.~\ref{table:loss_layer}, we could observe that $L1$ loss shows the best score compared to other two loss functions while `Cosine' shows the worst score which is even lower than `Pseudo-Only'.

We also test about on which layer of the segmentation network the self-attention loss would work the best. A segmentation network mainly consists of two parts, a feature extractor, $\mathcal{F}$, and a classification head, $\mathcal{H}$, hence $\mathcal{G}(x) = \mathcal{H}(\mathcal{F}(x))$. 'Prediction' and 'Feature map' in the Tab.~\ref{table:loss_layer}, refer to applying our proposed self-attention loss on $z$ and the feature map, $\mathcal{F}(x) = f \in \mathbb{R}^{h\times w\times k}$, where $k$ refers to the number of channels. `Both' is applying our method on both $z$ and $f$. We find that computing self-attention loss only on the prediction $z$ achieves the best performance. We conjecture this result is due to the fact that applying our method only on `Feature map' can not train the classification head and applying on both layers regularizes the network excessively more than necessary. The layer analysis experiments are done using the $L1$ loss.

\section{Selecting the $\tau^c$ for pseudo label generation}
$\tau^c$ is set differently for each class as mentioned in the main paper. We basically follow the pseudo label generation process of \cite{li2019bidirectional}. Pseudo labels are generated using a pre-trained segmentation network. The pre-trained network $\mathcal{G}$ makes inference on all images in the training set of the target domain to obtain the prediction results. Then, for each class, we collect all the pixels that are classified as the class from the entire predicted results and add the confidence score of each pixel to a list. Therefore, there is one list for each class. We sort each list and choose the median value of the list as the $\tau^c$ for the corresponding class, hence $\tau^c$ is set by the confidence score of top 50\% of each class. If the chosen median value is higher than 0.9, $\tau^c$ is set as 0.9.

\section{Discussion about why different settings work for different source domains}

\begin{figure*}[t]
    \centering
    \includegraphics[width = 0.9\linewidth]{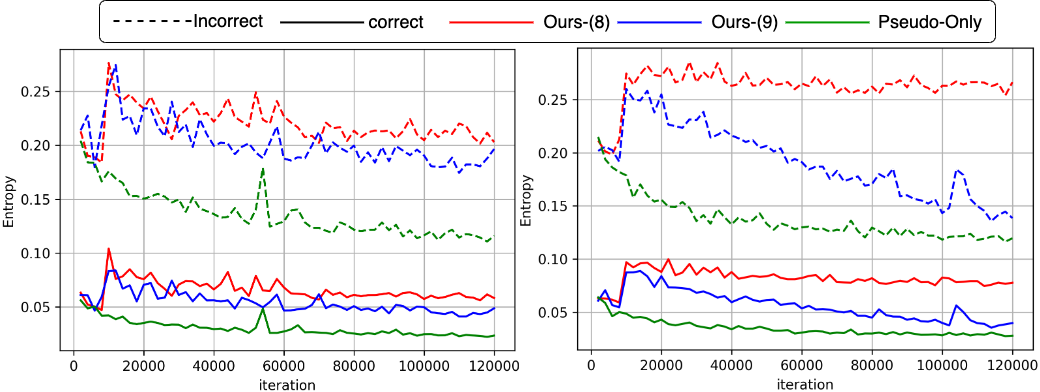}
    \vspace{-3mm}
    \caption{Entropy analysis of GTA5\textrightarrow Cityscapses (Left) and SYNTHIA\textrightarrow Cityscapses (right).}
    \vspace{-5mm}
    \label{fig:supp_quantitative_analysis}
\end{figure*}

From the Tab.~1 of the main paper, we could observe that different settings are suitable for different source domains. It is difficult to prove why this tendency happens exactly, but we can conjecture by analysis. Fig.~\ref{fig:supp_quantitative_analysis} is the same entropy analysis figure from the main paper (Fig.~\ref{fig:qualitative_analysis}). We think we can get a little hint from it. Ours (\ref{att_loss_1}) shows higher entropy than Ours (\ref{att_loss_2}) generally on both UDA tasks. However, for SYNTHIA\textrightarrow Cityscapses, Ours (\ref{att_loss_1}) show much higher entropy than Ours (\ref{att_loss_2}) compared to GTA5\textrightarrow Cityscapses. We can clearly observe the large gap between Ours (\ref{att_loss_1}) and Ours (\ref{att_loss_2}) on SYNTHIA task. This gap is much larger than that of GTA5 task, especially for incorrect pixels. Ours (\ref{att_loss_1}) is defined as minimizing the $L1$ loss between $z$ and $z''$, where $z'' = z + z'$. On SYNTHIA task, adding the output of the segmentation network, $z$ to the output of the SAM, $z'$ somehow brings noisy and incorrect information and eventually make $z''$ corrupted and under-perform. 

We guess this is because of the larger domain gap between SYNTHIA and Cityscapes than GTA5 and Cityscapes. In fact, SYNTHIA has very different data distribution from Cityscapses and GTA5. Cityscapses and GTA5 are both collected under driving scenario but SYNTHIA is not, it has more images taken from higher position such as traffic surveillance cameras. Also its number of classes shared with Cityscapses (16 classes) is less than GTA5 (19 classes). Therefore due to this larger domain gap, $z$ itself does not contain well-represented domain-invariant information that can improve the performance but rather deteriorates performance when combined with $z'$. On the other hand $z'$ which is refined version of $z$ with the help of the SAM, contains domain-invariant information than can further boost the performance. For this reason, we conjecture that it is better to just follow $z'$ instead of $z''$ for SYNTHIA task. 

We think the same reason applies to why using the self-attention loss only on the target domain works better than using it on both domains. If the self-attention loss is applied on both the source domain and the target domain, the network could be more overfitted and trained towards the source domain. This is not desirable especially when there is a large domain gap between the source and the target domains, such as SYNTHIA and Cityscapes.

\section{More Qualitative Results}

\subsection{Attention and Prediction visualization}
\begin{figure*}[t]
    \centering
    \includegraphics[width = 0.98\linewidth]{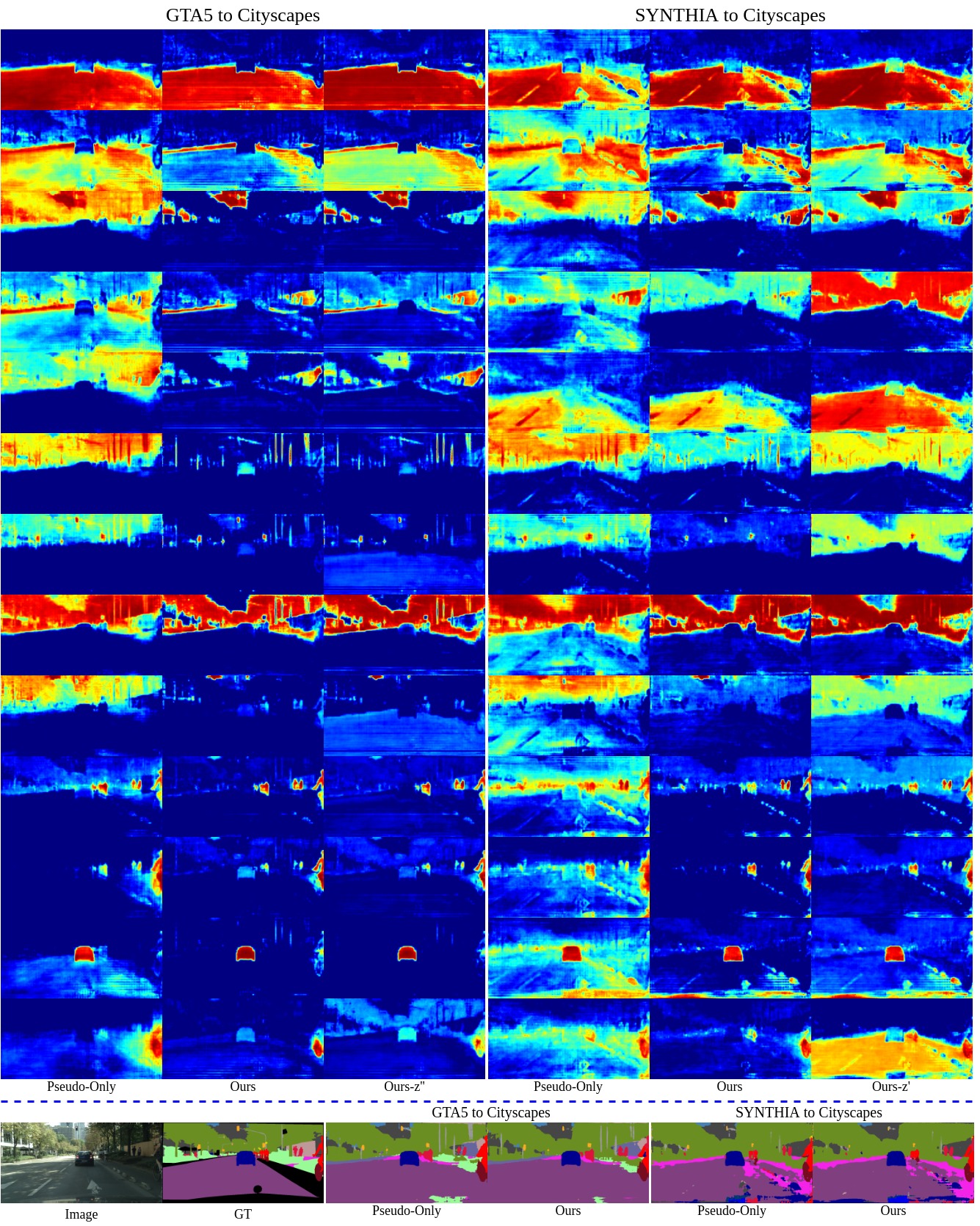}
    \caption{More attention and prediction visualization.}
    \label{fig:supp_qualitative_1}
\end{figure*}

\begin{figure*}[t]
    \centering
    \includegraphics[width = 0.98\linewidth]{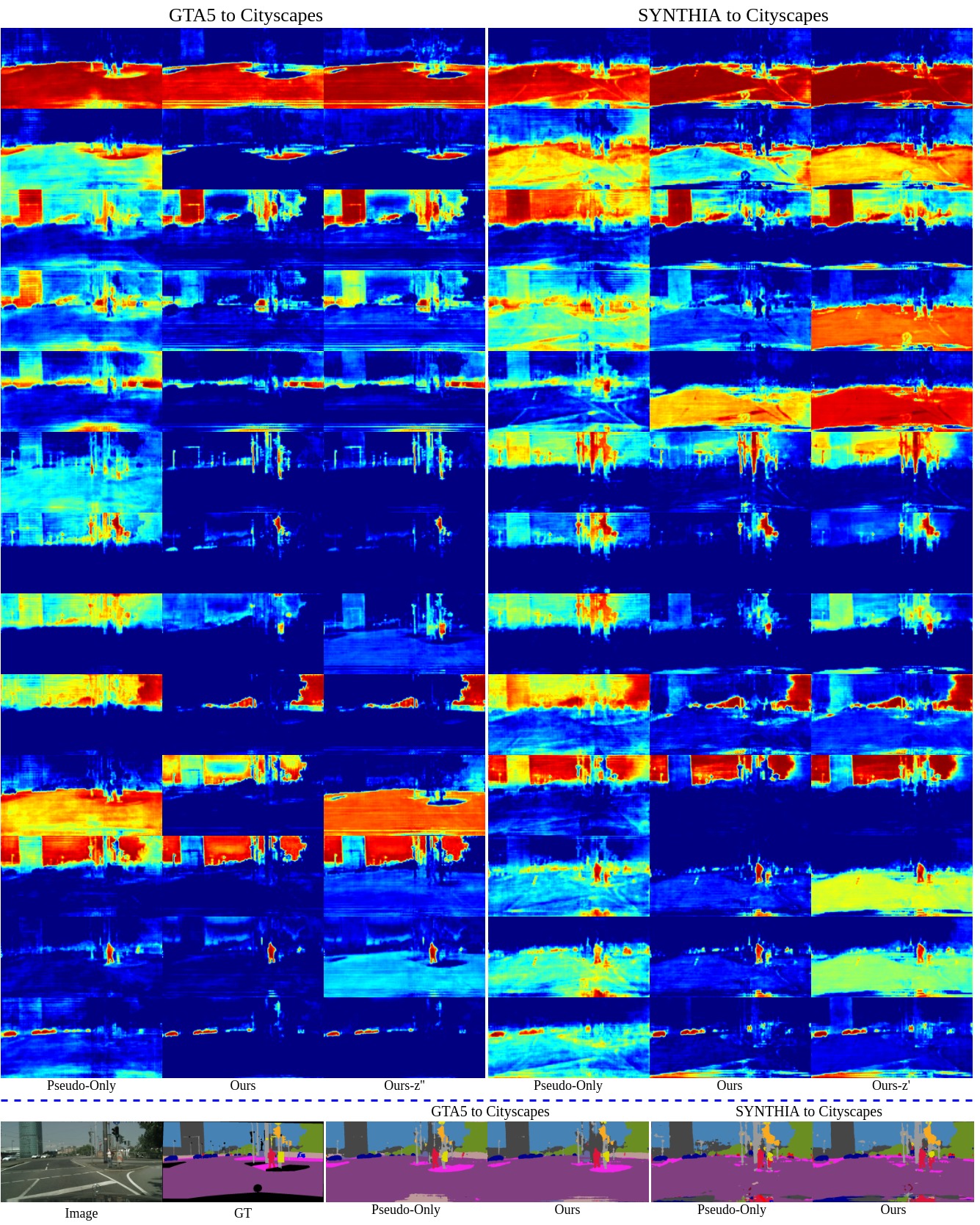}
    \caption{More attention and prediction visualization.}
    \label{fig:supp_qualitative_2}
\end{figure*}

\begin{figure*}[t]
    \centering
    \includegraphics[width = 0.98\linewidth]{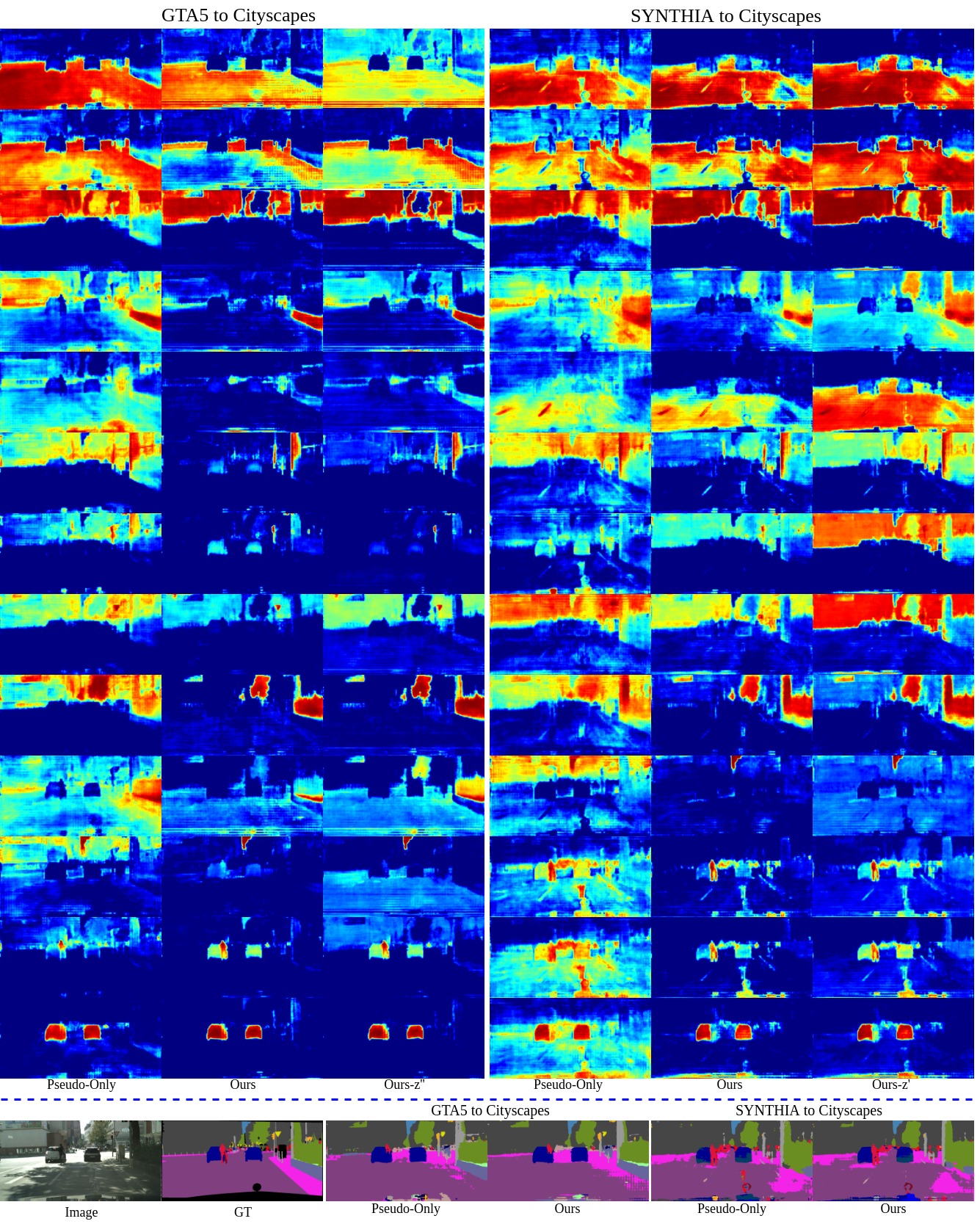}
    \caption{More attention and prediction visualization.}
    \label{fig:supp_qualitative_3}
\end{figure*}

Fig.~\ref{fig:supp_qualitative_1}-\ref{fig:supp_qualitative_3} show more qualitative results of attention and prediction visualization introduced in Sec.~\ref{qualitative} of main paper. Each figure shows the attention and prediction visualizations of an image on GTA5\textrightarrow Cityscapses and SYNTHIA\textrightarrow Cityscapses tasks. Each row of attention visualization refers to a different class. 

\subsection{Pixel-wise similarity visualization}
\begin{figure*}[t]
    \centering
    \includegraphics[width = 0.85\linewidth]{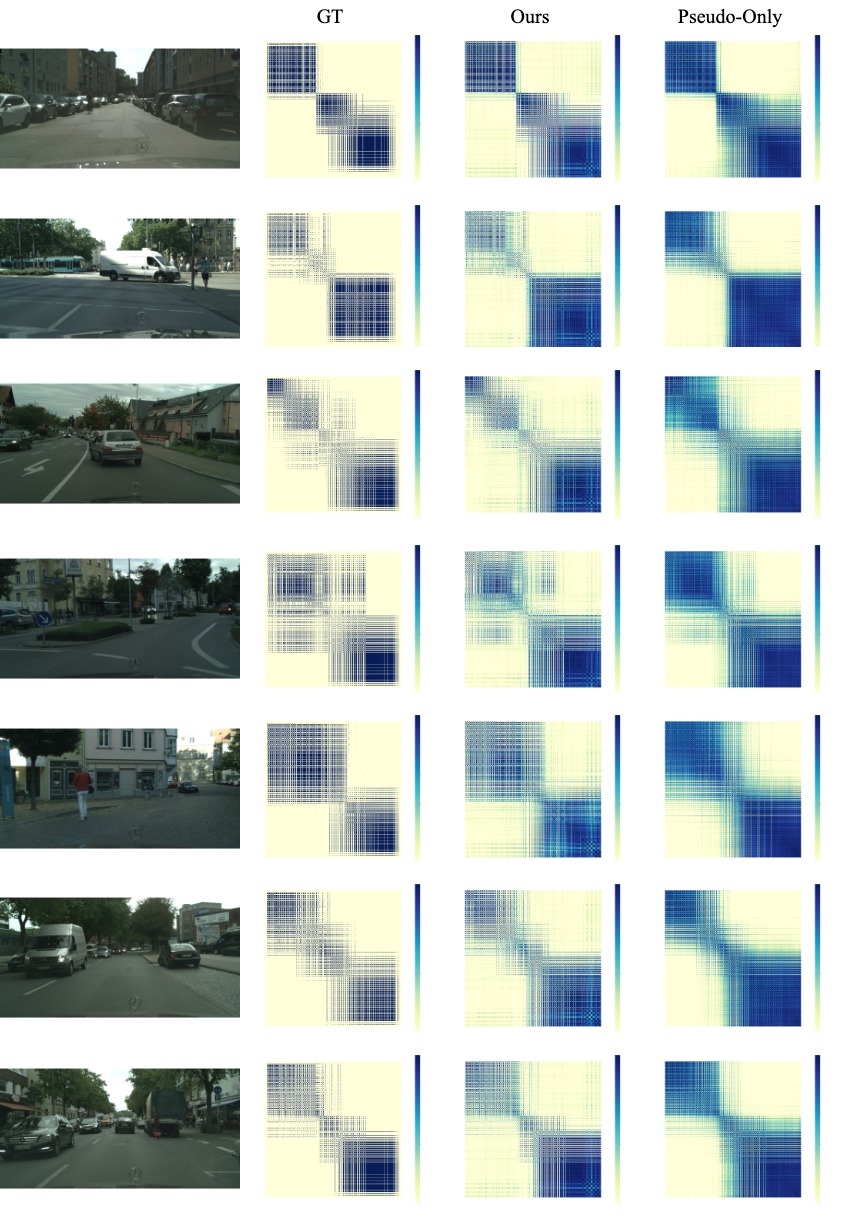}
    \vspace{-5mm}
    \caption{Comparison of pixel-wise similarity between `Ours' and `Pseudo-Only'.}
    \vspace{-5mm}
    \label{fig:supp_attention_map}
\end{figure*}

In Fig.~\ref{fig:supp_attention_map}, we show visualization of pixel-wise similarity. We visualize how each logit of predicted pixel ($z \in \mathbb{R}^{hw\times C}$) is similar to other pixels. It is computed as follows:
\begin{equation}
    M = ReLU(\frac{{z}\boldsymbol{\cdot} z^\mathsf{T}}{\norm{z}_2 \boldsymbol{\cdot} \norm{z}^\mathsf{T}_2}) \in \mathbb{R}^{hw\times hw}
    \label{eq:pixel-wise}
\end{equation}
It is basically an attention map of $z$ itself. We take ReLU on the attention map to visualize the difference in positive correlation between pixels more prominently. We visualize this for both `Ours' and `Pseudo-Only'.
For the ground truth, we use the nearest interpolation to resize the ground truth label to the spatial size of $h \times w$ from $H \times W$, hence $y_t \in \mathbb{R}^{h\times w\times C}$ where each pixel is a $C$ dimensional one-hot vector. We flatten $y_t$ in the spatial dimension ($y_t \in \mathbb{R}^{hw\times C}$) and compute $M$ by inserting $y_t$ instead of $z$ in (\ref{eq:pixel-wise}). 

The experiment is conducted using DeepLabV2 with ResNet101 backbone segmentation network that is trained on GTA5\textrightarrow Cityscapses task. We sample several images from validation set of Cityscapses and visualize the pixel-wise similarity described as above. In the figure, bluer means higher similarity. Since `GT' is visualized using the one-hot vectors of $y_t$, its each element is either 0 or 1 while elements of `Ours' and `Pseudo-Only' are between 0 and 1. 

As shown in the figure, `Ours' show much similar results to `GT' compared to 'Pseudo-Only' and clear difference between `Ours' and `Pseudo-Only' can be observed. It means that each pixel of $z$ which is a $C$ dimensional logit, is more attended well with the correct pixels corresponding to the same class and dissimilar to other pixels of different classes. On the other hand, pixels of `Pseudo-Only' are attended with even irrelevant pixels showing high similarity. 

\nocite{pan2020unsupervised}
\nocite{vu2019dada}
\nocite{zhao2020review}
\nocite{wang2020classes}

\newpage
{\small
\bibliographystyle{ieee_fullname}
\bibliography{egbib}
}

\end{document}